\documentclass[journal,twoside,web]{ieeecolor}
\usepackage{generic}
\usepackage{xcolor}
\usepackage{cite}
\usepackage{amsmath, amssymb, algorithm, algpseudocode}
\usepackage{graphicx}
\usepackage{algorithm}
\usepackage{hyperref}
\hypersetup{hidelinks=true}
\usepackage{textcomp}

\usepackage{multirow}
\usepackage{booktabs}

\usepackage{subfigure}
\usepackage{tabularx}
\usepackage{threeparttable} 
\usepackage{makecell}

\newcommand{\etal}{\emph{et al.}}
\newcommand{\eg}{\emph{e.g.}}
\newcommand{\ie}{\emph{i.e.}}
\usepackage{siunitx}
\def\BibTeX{{\rm B\kern-.05em{\sc i\kern-.025em b}\kern-.08em
    T\kern-.1667em\lower.7ex\hbox{E}\kern-.125emX}}
\begin{document}
\title{RIHA: Report-Image Hierarchical  Alignment for Radiology Report Generation}
\author{Yucheng Chen, Yang Yu, Yufei Shi, Conghao Xiong, Xulei Yang, and Si Yong Yeo
\thanks{©
2026 IEEE. The final published version is available at https://doi.org/10.1109/JBHI.2026.3670023.}
\thanks{This work is supported by the Lee Kong Chian School of Medicine (NTU) Start-Up Grant (award number: 025277-00018).}
\thanks{Yucheng Chen, Yufei Shi, and Si Yong Yeo are with MedVisAI Lab, Lee Kong Chian School of Medicine, Nanyang Technological University (NTU), and Centre of AI in Medicine, Singapore (email: yucheng005@e.ntu.edu.sg; yufei005@e.ntu.edu.sg; siyong.yeo@ntu.edu.sg). }
\thanks{Yang Yu and Xulei Yang are with Department of Machine Intellection, Institute for Infocomm Research (I$^{2}$R), Agency for Science, Technology and Research, (A*STAR), Singapore (email: yu\_yang@a-star.edu.sg; yang\_xulei@a-star.edu.sg).}
\thanks{Conghao Xiong is with the Department of Computer Science and Engineering, the Chinese University of Hong Kong, Hong Kong SAR, China (email: chxiong21@cse.cuhk.edu.hk).}
\thanks{Si Yong Yeo is the corresponding author. }
}

\maketitle

\begin{abstract}
Radiology report generation (RRG) has emerged as a promising approach to alleviate radiologists’ workload and reduce human errors by automatically generating diagnostic reports from medical images. A key challenge in RRG is achieving fine-grained alignment between complex visual features and the hierarchical structure of long-form radiology reports. Although recent methods have improved image-text representation learning, they often treat reports as flat sequences, overlooking their structured sections and semantic hierarchies. This simplification hinders precise cross-modal alignment and weakens RRG accuracy.
To address this challenge, we propose RIHA (Report-Image Hierarchical Alignment Transformer), a novel end-to-end framework that performs multi-level alignment between radiological images and their corresponding reports across paragraph, sentence, and word levels. This hierarchical alignment enables more precise cross-modal mapping, essential for capturing the nuanced semantics embedded in clinical narratives.
Specifically, RIHA introduces a Visual Feature Pyramid (VFP) to extract multi-scale visual features and a Text Feature Pyramid (TFP) to represent multi-granularity textual structures. These components are integrated through a Cross-modal Hierarchical Alignment (CHA) module, leveraging optimal transport to effectively align visual and textual features across various levels. Furthermore, we incorporate Relative Positional Encoding (RPE) into the decoder to model spatial and semantic relationships among tokens, enhancing the token-level alignment between visual features and generated text. 
Extensive experiments on two benchmark chest X-ray datasets, IU-Xray and MIMIC-CXR, demonstrate that RIHA outperforms existing state-of-the-art models in both natural language generation and clinical efficacy metrics.
\end{abstract}


\section{Introduction}
\label{sec:introduction}
\begin{figure*}[!t]
\centerline{\includegraphics[width=0.8\textwidth]{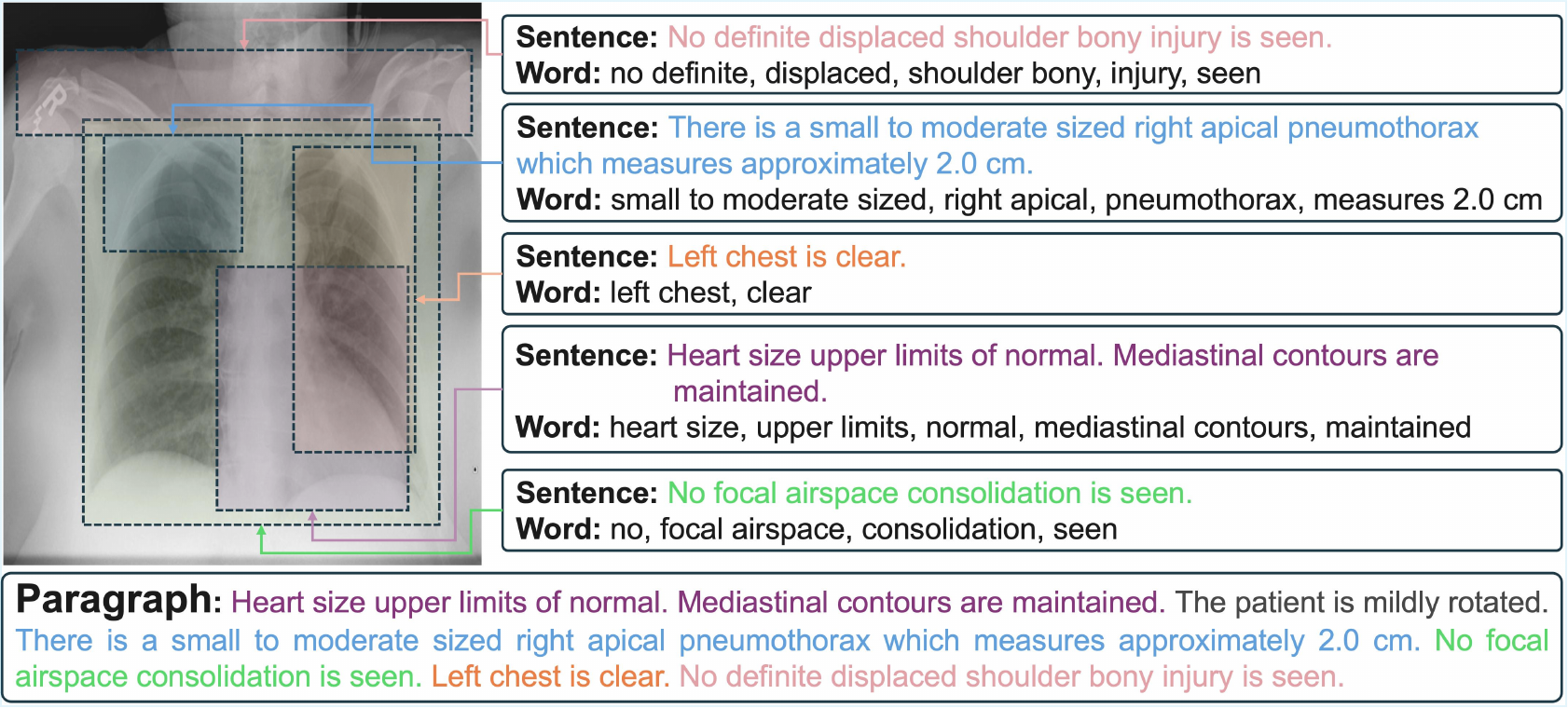}}
\caption{A chest X-ray image alongside its corresponding radiology report. The boxes on the right display, respectively, the individual sentences and the extracted keywords from the report. Multi-level visual-textual alignments are indicated using matching colours to highlight their associations.}
\label{fig1}
\vspace{-5pt}
\end{figure*}

\IEEEPARstart{R}{adiology} report generation (RRG) aims to produce accurate free-text descriptions of radiology images (e.g., chest X-rays) to support clinical decision-making and patient treatment. In clinical practice, interpreting radiographic images and composing diagnostic reports is a labor-intensive task that demands the specialized expertise of radiologists. This challenge is further compounded by a global shortage of radiology professionals~\cite{konstantinidis2024shortage}.  An exemplar radiology report with multiple granularities is also shown in Fig.~\ref{fig1}. Unlike natural images, radiological scans are characterized by high structural complexity, abstraction, and domain-specific visual patterns, making them more difficult to describe and interpret. As a result, establishing direct and well-defined structural correspondences between visual content and textual descriptions becomes particularly challenging~\cite{chen2023fine}. Thus, automated radiology report generation has emerged as a promising solution to reduce radiologists' workload while preserving the accuracy and quality of the reports~\cite{reale2024vision}.

Recent advancements in artificial intelligence have significantly propelled research in RRG~\cite{reale2024vision, hartsock2024vision}, offering potential to alleviate radiologists' workloads and address staffing shortages in hospitals~\cite{konstantinidis2024shortage}. While RRG is often framed similarly to image captioning~\cite{xu2015show}, substantial differences introduce unique challenges when applying conventional techniques to medical contexts~\cite{jing2017automatic}. Unlike image captioning that generates balanced descriptions of visual concepts, RRG emphasizes detecting abnormal findings within predominantly normal anatomical structures~\cite{li2024mlip}, and involves generating multi-sentence, paragraph-length reports with complex semantic relationships~\cite{chatterjee2018diverse}. These extended narratives often contain complex semantic relationships that pose additional challenges for models to capture fine-grained alignments between visual and textual modalities. Crucially, the hierarchical structure of radiology reports directly reflects the clinical diagnostic workflow, where radiologists systematically process images from global anatomical assessment to regional abnormality detection to precise quantitative measurements~\cite {irvin2019chexpert}. This diagnostic approach manifests in reports where paragraph-level impressions provide overall clinical context, sentence-level descriptions detail specific anatomical findings, and word-level terminology captures precise medical measurements. The inherent hierarchical nature of medical reasoning demands sophisticated visual-textual alignment mechanisms that can capture multi-level correspondences essential for clinically coherent report generation.

Existing approaches to address these alignment challenges broadly fall into two categories: image-report-only methods~\cite{wang2022cross, chen2020generating, wang2024camanet, zhang2025medunifier, chen2022cross} that rely solely on available visual and textual data, and augmented knowledge approaches~\cite{li2023dynamic, huang2023kiut, li2022cross, zhao2021automatic, zhang2020radiology,yang2023radiology, yang2022knowledge} that incorporate external medical domain knowledge. In image-report-only methods, current multi-granularity approaches primarily adopt sequential or auxiliary task-based paradigms that fail to achieve comprehensive cross-modal correspondence. Li~\etal~\cite{li2023unify} employ a three-stage sequential framework that processes the alignment phase independently. Liu~\etal~\cite{liu2024multi} rely on sentence-level contrastive learning without extending to comprehensive hierarchical alignment. These methods typically focus on single-modal granularity processing - either visual or textual hierarchies - missing the opportunity to establish synchronized cross-modal correspondences. Moreover, existing hierarchical approaches process different semantic levels sequentially or independently, failing to capture the comprehensive multi-level visual-textual mappings necessary for maintaining clinical coherence. This fundamental gap between existing alignment paradigms and the simultaneous multi-granularity nature of medical reasoning motivates the need for novel approaches that can concurrently establish correspondence across all semantic levels through unified cross-modal hierarchical alignment.

To address the challenges of aligning complex visual and textual information in radiological imaging for the report generation task, we propose a Report-Image Hierarchical Alignment Transformer (RIHA), a novel framework designed to perform hierarchical alignment between radiological images and associated reports at three textual levels: paragraph, sentence, and word. This multi-level alignment enables fine-grained cross-modal mapping, which is essential for capturing the nuanced yet crucial semantics of radiology reports. RIHA also leverages a pyramid-based structure to encode visual data into multi-scale hierarchical representations, thereby facilitating a more accurate and semantically consistent alignment with the multi-granularity textual inputs. This design is especially beneficial for radiological vision-language models, where conventional methods often struggle due to the inherent mismatch between textual semantics and visual structures.  The main contributions of this work are summarized as follows:

\begin{enumerate}

\item We propose a novel end-to-end Report-Image Hierarchical Alignment (RIHA) method for improving fine-grained alignment between paired X-ray images and long-text radiology reports. RIHA captures hierarchical correspondences at three levels of granularity: shallow (word-level), intermediate (sentence-level), and high (paragraph-level) features.

\item We design a Visual Feature Pyramid (VFP) and a Text Feature Pyramid (TFP) extractor to capture multi-granularity visual and textual features. These modules coordinate with a Cross-modal Hierarchical Alignment (CHA) module, leveraging optimal transport to align semantic information across visual and textual modalities effectively.

\item To enhance token-level alignment between visual features and generated reports, we also integrate a Relative Positional Encoding (RPE) strategy into the decoder. This approach allows the model to generate contextually accurate radiology descriptions that closely correspond to the underlying visual features by capturing both spatial and semantic token relationships, thus improving semantic consistency and spatial accuracy.

\item We conduct comprehensive experiments on two publicly available X-ray datasets to demonstrate the substantial performance enhancements of our method over existing approaches in the report generation task. Additionally, we validate the robustness and adaptability of our approach to practical scenarios through a series of detailed ablation studies.

\end{enumerate}

\section{Related works}
\subsection{Image Captioning}
Image captioning, as the task of automatically generating natural language descriptions for images, has become a prominent research topic at the intersection of computer vision~\cite{yeo2011level,yang2016cardiac,yang2017novel,liu2025effdnet} and natural language processing~\cite{lu2017knowing,anderson2018bottom}. The predominant approach in this field adopts an encoder-decoder framework, in which a visual encoder (typically based on convolutional neural networks or Vision Transformers) first extracts semantic features from the input image~\cite{vaswani2017attention}. These features are then fed into a language decoder, often implemented using recurrent neural networks or Transformer models, to generate coherent textual descriptions~\cite{santhanam2020context,chen2020generating}. While early models relied on standard supervised learning, subsequent efforts have introduced more advanced techniques, such as attention mechanisms~\cite{sirshar2022attention}, reinforcement learning~\cite{rennie2017self}, and adversarial training~\cite{dai2017towards}, to enhance the quality and relevance of generated captions. More recently, the field has seen a shift toward large-scale vision-and-language pretraining (VLP) models, which leverage vast image-text corpora to learn rich multimodal representations, achieving state-of-the-art (SOTA) performance across various benchmarks~\cite{li2022blip,li2020oscar,wang2021simvlm}.

\vspace{-5pt}
\subsection{Radiology Report Generation}
The automatic generation of clinically accurate radiology reports has become a pivotal area of research. Recent approaches to medical report generation can be broadly categorized into two types based on their training data requirements and architectural frameworks: image-report-only and knowledge-augmentation-based paradigms. The knowledge-augmentation-based paradigm enhances model training by incorporating external medical domain knowledge to address the limitations of purely data-driven approaches that may lack essential clinical insight. For instance, Yang~\etal~\cite{yang2023radiology, yang2022knowledge} propose to enhance radiology report generation by incorporating medical knowledge: one learns a medical knowledge base with multi-modal alignment, and the other incorporates both general and specific medical knowledge. ~\cite{li2023dynamic} and~\cite{huang2023kiut} all employ the knowledge graph as the external knowledge to help the report generation. Another augmentation strategy is to incorporate external reference reports from existing databases, known as retrieval-augmented methods like~\cite{hou2024radiographic,yang2025spatio}. While these methods can improve performance, they introduce substantial practical challenges, including the need for specialized medical annotations and complex pipeline engineering. Our method belongs to the image-report-only paradigm, which relies exclusively on paired image-report data. Approaches within this paradigm treat reports as flat sequences, employing conventional attention mechanisms~\cite{wang2018tienet} for visual-textual correspondence, memory networks~\cite{chen2020generating} to store prototypical report structures, auxilliary modules~\cite{li2024organ,tian2024diffusion,jin2024improving} for knowledge learning, or vision-language pre-training models~\cite{zhang2025medunifier} that learn broad cross-modal representations, all focusing on global alignment between entire images and complete reports. Recent works have begun to recognize the hierarchical nature of medical reports, with Jing~\etal~\cite{jing2017automatic} generating reports section by section without establishing cross-modal hierarchical correspondence. Some approaches leverage multi-granularity information through auxiliary tasks, such as Liu~\etal~\cite{liu2024multi}, who employ sentence-level contrastive learning, though these methods require additional annotations or focus on specific granularity levels rather than comprehensive hierarchical alignment. Wang~\etal~\cite{wang2022multi} focus on generalized medical representation learning across multiple domains using cross-modal alignment at different granularities, but their approach targets general representation learning rather than report generation specificity. Despite these advances, existing Image-Report-Only methods either process granularities sequentially, focus on single-modal hierarchies, or require auxiliary supervision, with none achieving simultaneous cross-modal hierarchical alignment across all semantic levels, creating a fundamental gap between current alignment paradigms and the hierarchical nature of clinical reasoning that our work addresses.

\begin{figure*}
    \centering 
    \includegraphics[width=0.95\textwidth]{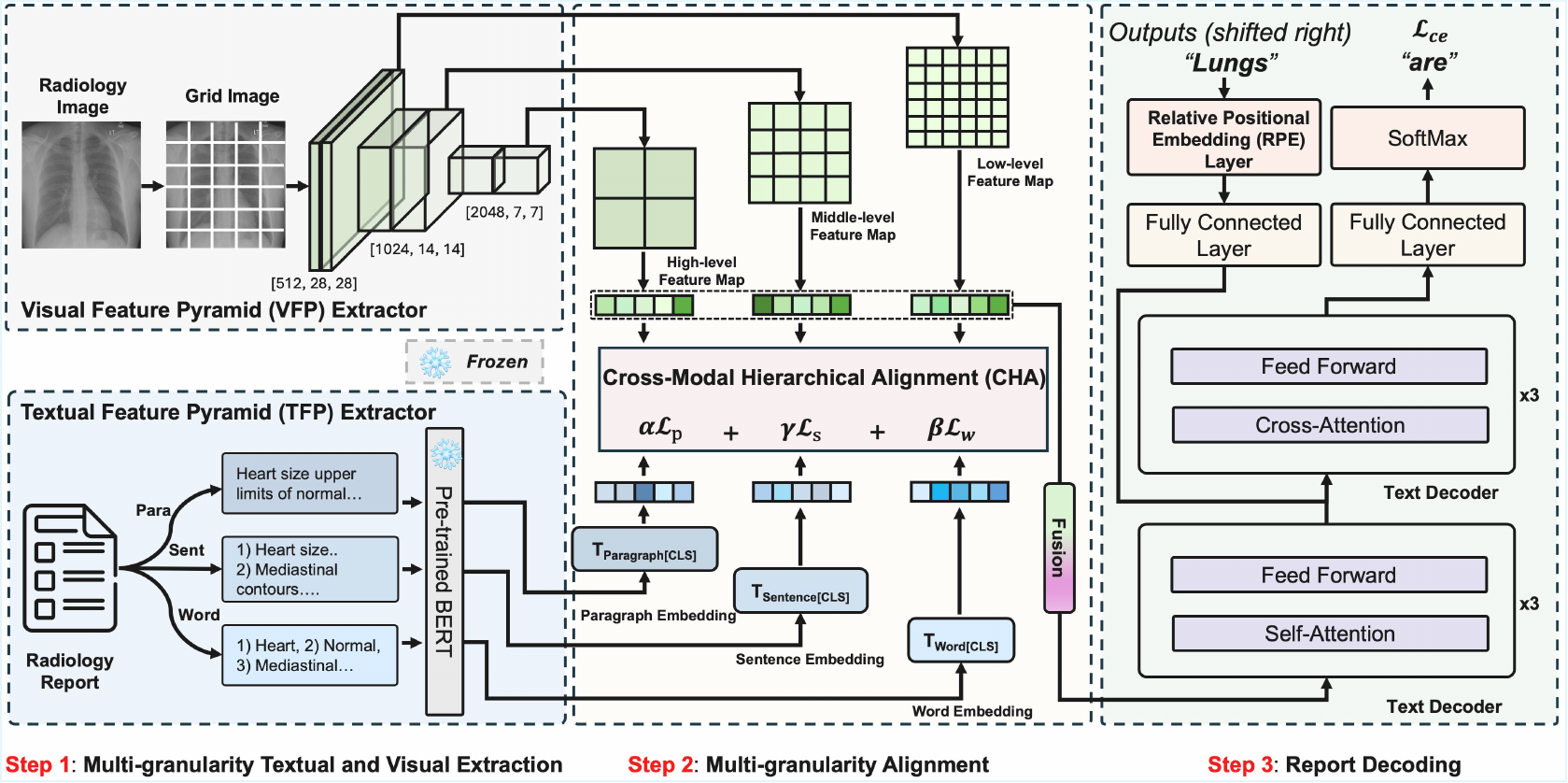} 
    \caption{The architecture of RIHA: An image is fed into the VFP Extractor to obtain shallow, middle, and high-level features. The multi-granularity text features of paragraph, sentence, and word-level features are extracted by the TFP extractor. Multi-granularity visual and textual features are then sent into CHA for hierarchical alignment. After that, refined visual and textual features are fed into a transformer encoder-decoder structure for report generation. }    
    \label{fig2}  
    \vspace{-10pt}
\end{figure*}

\vspace{-5pt}
\subsection{Cross-modal alignment}
A core challenge in multimodal learning lies in managing the inherent heterogeneity among diverse data modalities~\cite{zhu2024unraveling}. This heterogeneity is characterized by structural differences in data organization, distributional variations in feature spaces, and semantic disparities in information representation. Existing approaches fall into several major paradigms. The most prevalent approach to addressing modality heterogeneity is shared representation learning, which aims to map multiple modalities into a common latent space.  CLIP-based methods~\cite{gao2024clip} exemplify this strategy through contrastive learning on large-scale image-text pairs, while recent advancements such as Uni-Code~\cite{xia2023achieving} enhance alignment precision via cross-modal disentanglement. However, unifying modalities in a shared space often suppresses modality-specific characteristics. Transformer-based approaches address this through cross-attention mechanisms~\cite{tsai2019multimodal}, with techniques like hierarchical attention networks~\cite{yang2022disentangled} improving semantic granularity. Yet these methods typically demand substantial computational resources and struggle with highly disparate modality pairs. Modality translation represents an alternative direction, where explicit mappings between modalities are learned through cross-modal generation~\cite{liu2024modality}, cycle-consistency constraints~\cite{tian2022multimodal}, or hybrid encoder-decoder architectures~\cite{zeng2024disentanglement}. While effective for modality conversion, these methods risk introducing artifacts when handling complex, high-dimensional data. Cross-modal knowledge distillation has also emerged to address modality imbalance through dynamic knowledge transfer~\cite{li2023decoupled} and unified self-distillation frameworks~\cite{li2024unified}, enhancing cross-modal coherence while often relying heavily on teacher model quality. Across these paradigms, a consistent dilemma persists: methods either prioritize cross-modal alignment at the expense of modality specificity, or preserve modality characteristics while sacrificing integration effectiveness. To address this dilemma, we propose a unified framework that achieves multi-dimensional alignment from the perspective of distribution. By conducting alignment at the distribution level, our approach simultaneously preserves modality-specific information while achieving robust cross-modal integration.

\section{Method}
In this section, we present the RIHA model. Section \ref{sec:overview} provides an overview of radiology report generation, covering its mathematical formulation, basic network structure, and distance calculation method. The detailed model architecture, including each designed module, is discussed in Section \ref{sec:method}. Finally, the learning objectives are defined in Section \ref{sec:obj}.

\subsection{Overview and Preliminaries}
\label{sec:overview}

\textbf{Radiology Report Generation.}
Let us define $I$ as a radiology image, and the goal of RRG is to generate a report $R$ from $I$. Existing SOTA methods mainly utilize the encoder-decoder structure to generate reports. In practice, the visual encoder (\eg ResNet101) takes the X-ray images to extract visual features $F_{I} \in \mathbb{R}^{H \times W \times C}$, and then these features are flattened into a sequence of visual tokens $\hat{F}_{I} \in \mathbb{R}^{HW \times C}$, where the $H, W, C$ denotes the hight, width and the number of channels, respectively. The process can be formulated as follows,
\begin{equation}
\label{eq:vt}
\mathcal{F}_{ve}(I) = \{ \hat{f}_{1}^{i}, \hat{f}_{2}^{i},...,\hat{f}_{p}^{i},...,\hat{f}_{N}^{i} \}
\end{equation}
\noindent where $\hat{f}_{p}^{i}$ represents the $p$-th patch visual feature in $\hat{F}_{I}$, and $N=H\times W$. The visual extractor is denoted as $\mathcal{F}_{ve}$

After obtaining these visual features, they are sent to the transformer encoder-decoder framework to generate a report $R$. The generation is an auto-regressive procedure. In particular, we can obtain the word embeddings $e^{w} \in \mathbb{R}^{1 \times D}$ via an embedding layer for each word $w$ in Report. The visual tokens are transferred into intermediate representations $e^{i} \in \mathbb{R}^{1 \times D}$, where $D$ indicates the dimensions of hidden states. 
Following that, the word embeddings before time stamp $T$ combined with visual embeddings are input into the decoder for the generation of the word at the time stamp $T$. The whole process can be expressed in equations~\ref{eq:encoder},~\ref{eq:embedding} and ~\ref{eq:decoder}.

\textbf{Wasserstein Distance.} 
\label{definition}
The Wasserstein distance, which is also referred to as Earth Mover's Distance (EMD)~\cite{frogner2015learning, yang2023improving}, measures the distance between two probability distributions or collections of weighted objects. This metric is derived from the solution to the classical transportation problem in operations research~\cite{hitchcock1941distribution}. In mathematical terms, consider a system consisting of $n$ suppliers $S=\{s_{1}, s_{2}, ... , s_{n}\}$, where each element $s_{i}$ represents the supply capacity of the $i$-th source. These suppliers must distribute resources to $m$ customers with fixed demand quantities $Q= \{ q_{1}, q_{2},...,q_{m}\}$, where $ q_{j}$ denotes the requirement of the $j$-th consumer. The transportation cost per unit from supplier $i$ to customer $j$ is defined as $c_{ij}$. The optimization objective involves determining an optimal transportation plan $ P = \{p_{ij}| I$ $\in \lbrack 1,n \rbrack, j \in \lbrack 1, m \rbrack\} $ that minimizes total cost while satisfying all consumer demands. The formulations can be found in equation~\ref{eq2}.
\begin{align}
\label{eq2}
\underset{p_{ij}}{\text{minimize}} \quad & \sum_{i=1}^{n}\sum_{j=1}^{m}c_{ij}p_{ij} \nonumber\\
\text{subject to} \quad & p_{ij} \geq 0, \quad i = 1, \ldots, n, \, j = 1, \ldots, m \\
& \sum_{j=1}^{m}p_{ij} = s_i, \quad i = 1, \ldots, n \nonumber\\
& \sum_{i=1}^{n}p_{ij} = q_j, \quad j = 1, \ldots, m. \nonumber
\end{align}
To enhance the fine-grained alignment between X-ray images and their corresponding radiology reports, RIHA incorporates the Wasserstein distance as the fundamental component for cross-modal matching. Different from simple similarity measures~\cite{lahitani2016cosine,kullback1951kullback,rolle2021various,dokmanic2015euclidean}, Wasserstein distance provides a flexible, distribution-aware metric that captures the optimal transport cost between heterogeneous feature spaces. Radiological visual features extracted from different regions of the X-ray and semantic embeddings derived from text, spanning from word-, sentence-, and paragraph-hierarchies, are treated as probability distributions. By measuring the Wasserstein distance between these hierarchically encoded representations, the model learns to align visual and textual content at multiple granularities, enabling robust and semantically meaningful correspondences across modalities. This approach is particularly effective in the interpretation of long, descriptive radiology reports, where clinical observations are often distributed sporadically throughout the radiology report. To compute the optimal transportation plan $P$, we employ approximate algorithms~\cite{cuturi2013lightspeed, fan2017point, jonker1988shortest}. In this work, we leverage the Wasserstein distance to align the reports and images.

\begin{equation}
\label{eq:encoder}
   \mathcal{F}_{en}(\hat{f}_{1}^{i}, \hat{f}_{2}^{i},..., \hat{f}_{N}^{i}) = (e^{i}_{1}, e^{i}_{2}, ..., e^{i}_{N}),
\end{equation}
\vspace{-0.5cm}
\begin{equation}
\label{eq:embedding}
    \mathcal{F}_{em}(w_1, w_2,..., w_{T-1}) = (e^{w}_{1}, e^{w}_{2}, ..., e^{w}_{T-1}),
\end{equation}
\vspace{-0.5cm}
\begin{equation}
\label{eq:decoder}
  p_T = \mathcal{F}_{de}(e^{i}_{1}, e^{i}_{2}, ..., e^{i}_{N};e^{w}_{1}, e^{w}_{2}, ..., e^{w}_{T-1}),
\end{equation}
\noindent where $\mathcal{F}_{en}, \mathcal{F}_{em}$ and $\mathcal{F}_{de}$ represent the visual and embedding layer, and decoder, respectively. $p_{T}$ is the predicted word at time stamp $T$.

\subsection{Model Architecture}
\label{sec:method}
The overall framework of our proposed RIHA is illustrated in Fig. \ref{fig2}. It comprises three main stages: multi-granularity encoding, cross-modal alignment, and report decoding. In the first stage, X-ray images are processed through the visual feature pyramid (VFP) extractor to capture multi-level features. Simultaneously, the text feature pyramid (TFP) extractor encodes the input text at paragraph, sentence, and word levels. In the second stage, these multi-granularity textual features are then aligned with the corresponding visual features using the cross-modal hierarchical alignment (CHA) module. Notably, the CHA module is only required during training to learn hierarchical correspondences and can be removed during inference, significantly reducing computational overhead while maintaining the learned alignment benefits. Finally, for the third stage, the decoder incorporates the relative positional embedding (RPE) strategy to refine token-level alignment, enhancing overall coherence.

\textbf{Step 1: Multi-granularity Visual and Textual Feature Encoding.} 
Unlike conventional approaches that rely on limited textual and visual semantics, our method extracts multi-hierarchy visual features and disentangles long text into three levels of semantics: \ie paragraph, sentence, and word. Drawing inspiration from the pyramidal feature hierarchy, we capture multi-scale feature maps from various layers of the visual extractor to enhance the representation of visual content. The X-ray images are fed into a multi-hierarchical visual extractor VFP with a different number of patches, the higher the larger, to obtain shallow, middle, and high visual grid features denoted as $(\hat{V}_{s}, \hat{V}_{m}, \hat{V}_{h})$ in module VFP shown as blue component in Fig.~\ref{fig2}, where $\{s, m, h\} = \{5, 6, 7\}$ in ResNet and the $\hat{V}_{s}$ captures the feature from shallow layers while $\hat{V}_{h}$ represents the feature from deeper layer. For textual features, effectively encoding long text remains challenging despite its richer semantics. To address this, we employ TFP to decompose the long-text report into multiple levels of granularity: paragraph, sentence, and word. For sentence-level processing, we use NLTK's~\cite{bird2009natural} sentence tokenizer to split paragraphs into individual sentences, then filter out short sentences that contain minimal semantic information and prioritize sentences containing clinical findings over background information. For word-level encoding, we employ a systematic selection process rather than processing all words indiscriminately. Specifically, we use NLTK's POS tagger to identify parts of speech and selectively extract nouns, adjectives, and quantifiers that carry the most diagnostic information in radiology reports. We further prioritize medical terminology using clinical vocabulary filters and extract key phrases commonly appearing in radiology findings (e.g., ``consolidation," ``effusion," ``pneumothorax"). This selective approach ensures focus on clinically relevant semantic units.

Finally, a pre-trained Bio\_ClinicalBERT language model is applied to encode the resulting segments, with the model frozen during training to preserve medical domain knowledge. We obtain paragraph-level semantics directly from BERT embeddings, denoted as $t_{p} = \{ p_{1}, p_{2}, ..., p_{|t_{p}|}\}$, where $p_{i} \in \mathbb{R}^{d}$ represents the $i$-th word representation with dimension $d$. For sentence and word semantics, we extract representations from the $<$CLS$>$ token, which aggregates contextual information from all tokens through self-attention mechanisms, effectively capturing sequence-level semantics. These are denoted as $t_{s} = \{ s_{1}, s_{2}, ..., s_{|t_{s}|}\}$ and $t_{w} = \{w_{1}, w_{2}, ..., w_{|t_{w}|}\}$, where $s_{i}$ and $w_{i}$ indicate the $i$-th sentence and word representation with dimension $d$, respectively. For word-level encoding, $<$CLS$>$ representations are extracted from sequences containing selected keywords with minimal context to maintain semantic coherence.

\begin{figure}
    \centering 
    \includegraphics[width=0.4\textwidth]{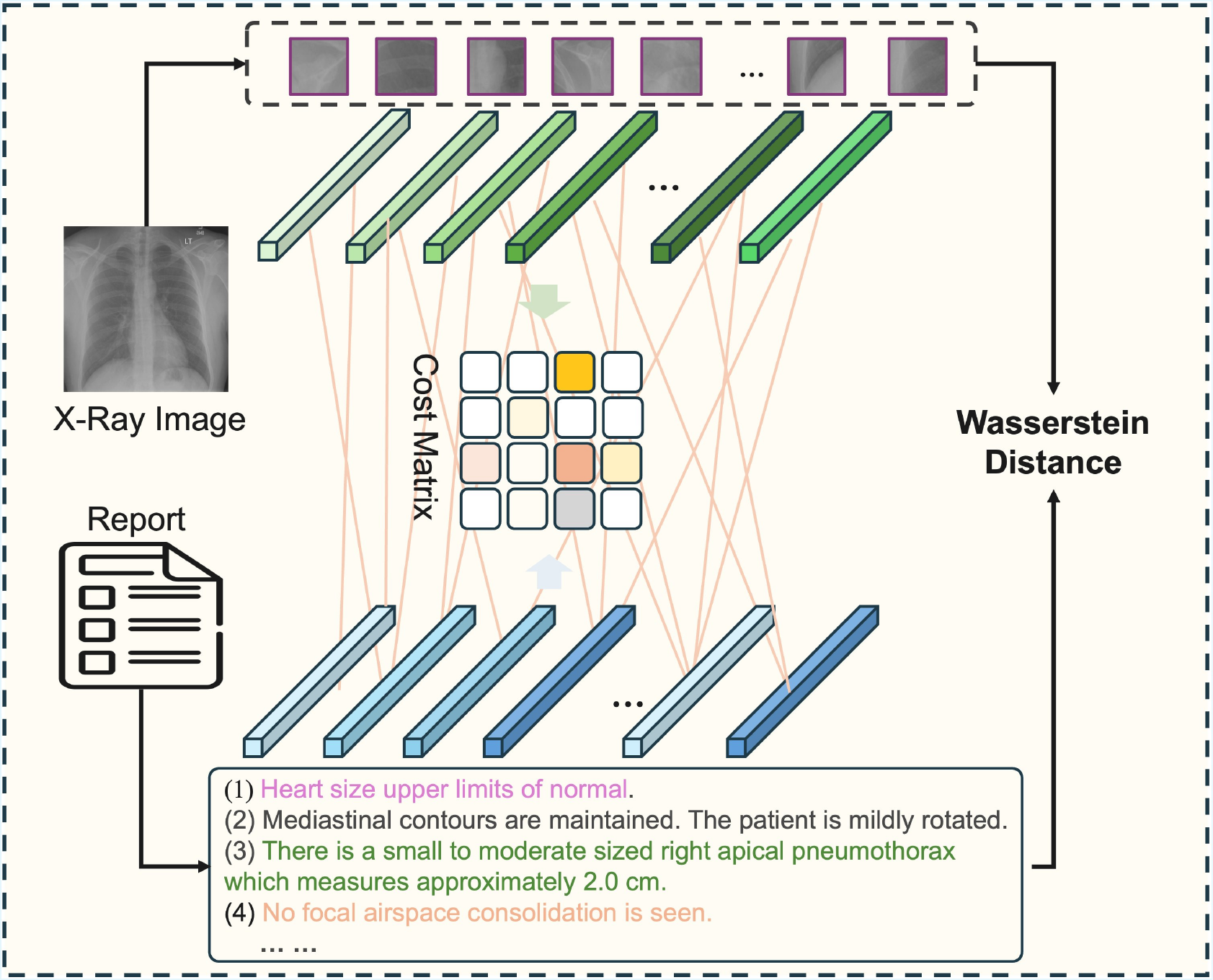} 
    \caption{An example of leveraging optimal transport to align hierarchical textual and visual features treats the grid visual features as suppliers and textual features as customers. The goal is to minimize the transportation cost, yielding an optimal alignment captured in the cost matrix.}   
    \label{fig:ot}  
\end{figure}

\textbf{Step 2: Multi-granularity Alignment.}
\label{calculation}
To address the alignment inconsistency between text and image, multi-granularity visual and textual semantics are first extracted as described in Step 1. However, effectively aligning these cross-modal features remains a critical challenge. Traditional methods often rely on mean representation distances, which overlook important local semantic cues. To overcome this, we model the alignment as an optimal transportation problem~\cite{villani2008optimal}, aiming to minimize the transportation cost from grid-based visual features to their corresponding textual representations. Specifically, we employ the Wasserstein distance to compute this cost, capturing the fine-grained correspondence between features. For illustration, we consider the alignment between a sentence $t_{s}$ and a middle-level visual feature $\hat{V}_{m}$ (see Fig.\ref{fig:ot}), as the alignment strategy remains consistent across different granularities. According to the definition in Sec.~\ref{definition}, we take $\hat{V}_{m}$ as the suppliers and $t_{s}$ as the customers. The transportation cost per unit can be calculated as the distance between the $i$-th visual grid representation and $j$-th sentence representation $l_{ij} = ||v_{i} - s_{j}||_{2}$. 
By minimizing this cost, we obtain the optimal transportation flow 
F, which can be integrated into the neural network for end-to-end training, enabling effective alignment across different semantic levels. After hierarchical alignment, the visual features from three hierarchies are fused for Step 3. Specifically, the aligned paragraph-level, sentence-level, and word-level visual features - $\hat{V}_h, \hat{V}_m, \hat{V}_s$ - are concatenated as $F_{fused}$ and then processed through a linear projection layer to reduce dimensionality before being fed into the transformer encoder.

\begin{figure}
    \centering 
    \includegraphics[width=0.4\textwidth]{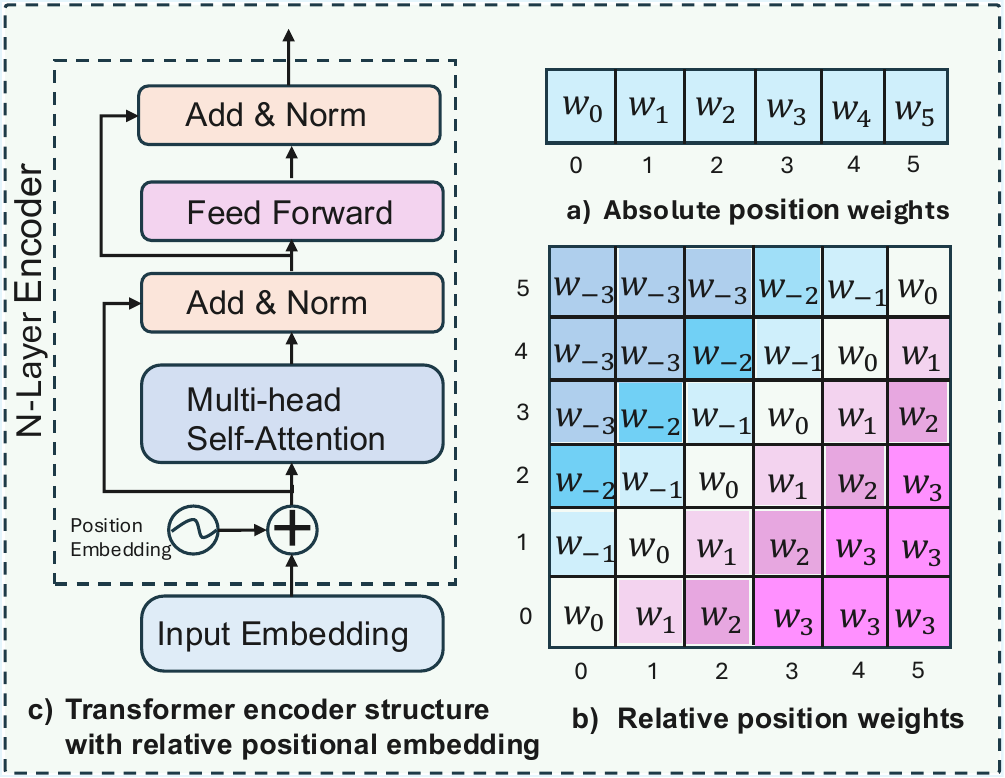} 
    \caption{An illustration comparing relative and absolute positional embeddings in transformers, where the clipped value $k=3$ represents the maximum allowable relative position distance. a) Absolute position embedding weights. b) Relative position embedding weights. c) The transformer encoder structure with relative position embeddings. For further details, see~\cite{wu2021rethinking}.}  
    \label{fig:rpe}  
    \vspace{-5pt}
\end{figure}

\textbf{Step 3: Report Decoding with Relative Position.}
In step 2, the hierarchical alignment process extracts enhanced visual features, which are then processed by the transformer encoder. This encoder utilizes multiple layers of multi-head self-attention and feedforward networks to capture complex visual representations. The decoder subsequently generates the radiology report, attending to both the previously generated tokens through masked self-attention and the encoded visual features via cross-attention mechanisms. To strengthen the alignment between the radiology report $R$ and the corresponding image $I$ at the token level, we incorporate relative position encoding (RPE) into the self-attention mechanism. Unlike the fixed positional encoding used in standard Transformers, which only captures absolute token positions, RPE effectively encodes relative positional information, directly modeling the contextual relationships between tokens as shown in Fig~\ref{fig:rpe}. This approach addresses the limitations of vanilla Transformers, which struggle with cross-modal alignment due to their reliance on fixed, sequence-based indices. The relative position representation and more details can be referred to~\cite{shaw2018self}.

\vspace{-5pt}
\subsection{Learning Objectives}
\label{sec:obj}
The overall objective function of our model includes CrossEntropy loss $\mathcal{L}_{ce}$ and hierarchical alignment loss $\mathcal{L}_{alignment}$. According to the calculation in Sec.~\ref{calculation}, the hierarchical alignment loss could be further separated into paragraph loss $\mathcal{L}_{p}$ (\ie $V_{h}$ and paragraph $t_{p}$), sentence loss $\mathcal{L}_{s}$ (\ie $V_{m}$ and sentence $t_{s}$), and word loss $\mathcal{L}_{w}$ (\ie $V_{w}$ and word $t_{w}$). We introduce the weights $\alpha, \gamma, \beta$ to balance the contributions of these components. Therefore, the overall learning objectives for the proposed model are as follows.

\begin{equation}
\label{eq:sim}
\mathcal{L}_{alignment} = \alpha\mathcal{L}_{p} + \gamma\mathcal{L}_{s} + \beta\mathcal{L}_{w},
\end{equation}
\begin{equation}
\mathcal{L} = \mathcal{L}_{ce} + \mathcal{L}_{alignment}
\end{equation}

\section{Experiments}
\label{sec:exp}
\subsection{Datasets and Evaluation Metrics}
\textbf{IU-Xray}\cite{demner2016preparing} is a widely used chest X-ray dataset containing 3,996 reports from the Indiana Network and 8,121 associated frontal or lateral view images from hospital picture archiving systems. Each report is composed of three sections: indication, findings, and impression. In practical experimental settings~\cite{chen2020generating, li2018hybrid, liu2021exploring}, only the findings section is used as the target text for generation based on the paired images. Reports lacking a findings section are excluded, reducing the actual dataset to 2,955 reports. For training, the dataset is typically split into 2,069 training, 296 validation, and 590 testing samples, with no overlap between subsets, as each report corresponds to a unique patient. Additionally, previous studies have standardized the maximum report length to 60 tokens.

\textbf{MIMIC-CXR}~\cite{johnson2019mimic} is a comprehensive dataset comprising 227,835 imaging studies from 65,379 patients, collected at the Beth Israel Deaconess Medical Center Emergency Department between 2011 and 2016. Each report includes sections such as Indication, Preamble, Comparison, Findings, and Impression. In this work, we focus on generating the Findings section as the RRG task. The dataset provides 473,057 X-ray images and 206,563 corresponding reports. For consistency with prior studies, we adopt the official data split: 152,173 reports and 270,790 images for training, 1,196 reports and 2,130 images for validation, and 2,347 reports and 3,858 images for testing. Unlike the IU-Xray dataset, we use single X-ray images paired with their respective reports, as approximately 35\% of the MIMIC-CXR reports are associated with only one image. The maximum report length is set to 100 tokens.

\textbf{Evaluation Metrics} for RRG encompass two main aspects: natural language generation (NLG) and clinical efficacy (CE). NLG performance is measured using BLEU~\cite{papineni2002bleu}, METEOR~\cite{denkowski2011meteor}, ROUGE-L~\cite{lin2004rouge}, and CIDEr-D~\cite{vedantam2015cider}. Clinical efficacy is assessed via the CheXpert tool~\cite{irvin2019chexpert}, which labels each report based on the presence of 14 findings, where $1$, $0$, -$1$, and \textit{blank} denote confidently present, confidently absent, uncertainly present, and not mentioned, respectively. Consistent with prior studies, we classify $1$ and -$1$ as positive labels, while $0$ and \textit{blank} are considered negative. We then compute the mean precision, recall, and F1 score across the 14 classes using these labels from both the predicted and ground truth reports.
\vspace{-5pt}
\subsection{Implementation Details}
The model is trained end-to-end using a PyTorch implementation on a single NVIDIA A6000 GPU. Before feeding images into the model, we crop the images into $224 \times 224$ pixels, and use the ResNet101 pretrained on ImageNet as the backbone of the visual extractor. The final output from the visual extractor is a $7 \times 7$ feature map with the dimension of each feature set to 2048. For extracting the hierarchical textual feature, we first leverage the natural language toolkit (\ie NLTK) to split the paragraph into sentence- and word-level information and save it to independent configuration files. Then, we fed three levels of textual content into a pre-trained BERT model, where we employ the Bio\_ClinicalBERT, to extract the hierarchical textual features. For the encoder-decoder backbone, we employ a randomly initialized memory-driven Transformer~\cite{chen2020generating} with 3 layers and 8 attention heads, and 512 dimensions for the hidden states. The hyperparameter settings in Wasserstein distance, including the regularization coefficient $\sigma$ and the marginal penalization $\tau$, are set as $0.1$ and $0.5$, respectively. The value for each hyperparameter depends on the performance of the model on the validation subset.
During the training stage, the learning rate is set as $1e-3$ for the visual extractor and $2e-3$ for the encoder-decoder on the IU-Xray dataset, while the values are $5e-5$ and $1e-4$ on the MIMIC-CXR dataset. We use Adam to optimize our model. In addition, the training epoch and batch size are set as $30$ and $64$, respectively.
\vspace{-5pt}
\subsection{Comparison to SOTA Methods}
In this section, we compare our method with a wide range of methods, including the conventional image captioning~\cite{lu2017knowing,rennie2017self} as well as previous RRG approaches~\cite{li2018hybrid,chen2020generating,liu2022competence,chen2022cross, liu2021exploring, tanida2023interactive, wang2022cross, wang2024camanet, jin2024promptmrg}. Most of the methods evaluate their performance on the two public datasets - IU-Xray and MIMIC-CXR - and share similar experimental settings as those used in work~\cite{chen2020generating}, so we cite some experimental results directly from their papers. In the comparisons, ATT2IN~\cite{rennie2017self} applies a reinforcement learning strategy originally designed for image captioning to report generation. R2Gen~\cite{chen2020generating} introduces a relational memory mechanism to capture key information for automatic radiology report generation. R2GenCMN~\cite{chen2022cross} further advances this approach with a cross-modal memory network, enhancing multimodal interactions. XProNet~\cite{wang2022cross} constructs a cross-modal prototype network to strengthen cross-modal pattern learning, thereby improving report accuracy. RGRG~\cite{tanida2023interactive} adopts an interactive, explainable approach by detecting anatomical regions and generating localized descriptions. M2KT~\cite{yang2023radiology} and GSKET~\cite{yang2022knowledge} enhance radiology report generation by incorporating medical knowledge: the former learns a medical knowledge base while performing multi-modal alignment, and the latter incorporates both general medical knowledge and disease-specific knowledge. CAMANET~\cite{wang2024camanet} employs class activation maps to guide attention mechanisms for generating more accurate and visually-grounded radiology reports from medical images. Finally, PromptMRG~\cite{jin2024promptmrg} leverages diagnosis-driven prompts, transforming disease classification results into token prompts that guide the report generation process.

As shown in Tab~\ref{tab:main_result}, our proposed RIHA method outperforms most of the state-of-art approaches on two benchmark datasets. For the IU-Xray dataset, RIHA achieves superior performance across multiple metrics, obtaining the highest scores in BLEU-1 (0.498), BLEU-4 (0.218), METEOR (0.204), and CIDEr (0.421). Notably, our method outperforms recent strong baselines including M2KT~\cite{yang2023radiology} by 1.8\% on BLEU-1 and 25.3\% on BLEU-4, GSKET~\cite{yang2022knowledge} by 1.6\% on BLEU-1 and 23.2\% on BLEU-4, CAMANET~\cite{wang2024camanet} by 4.2\% on BLEU-1 and 21.8\% on BLEU-4 and PromptMRG~\cite{jin2024promptmrg} by 2.0\% on BLEU-1 and 15.3\% on BLEU-4.
On the more challenging MIMIC-CXR dataset, RIHA demonstrates competitive results, achieving the best scores in BLEU-1 (0.385), BLEU-2 (0.251), METEOR (0.159), and ROUGE-L (0.293). Our method shows consistent improvements over recent approaches including M2KT~\cite{yang2023radiology} (4.6\% improvement on BLEU-1), CAMANET~\cite{wang2024camanet} (4.9\% improvement on BLEU-1), and COMG~\cite{gu2024complex} (11.6\% improvement on BLEU-1).
These results demonstrate the effectiveness of our proposed method in generating accurate and semantically meaningful radiology reports across different datasets.

\begin{table*}[ht]
\footnotesize
\setlength{\fboxsep}{0.8pt}
\begin{center}
\caption{Comparison with state-of-the-art methods on IU-Xray and MIMIC-CXR datasets. \colorbox{blue!15}{Optimal} and \colorbox{green!15}{suboptimal} performance is highlighted. }
\label{tab:main_result}
\begin{tabular}{@{}l l c c c c c c c@{}}
\toprule
\textsc{\textbf{Dataset}}& \textsc{\textbf{Method}} & \textbf{BLEU-1} & \textbf{BLEU-2} & \textbf{BLEU-3} & \textbf{BLEU-4} & \textbf{METEOR}  & \textbf{ROUGE-L} & \textbf{\textsc{CIDEr}} \\
\midrule 
\multirow{11}{*}[-3pt]{IU-Xray~\cite{demner2016preparing}} & ADDAATT~\cite{lu2017knowing}  & $0.220$ & $0.127$ & $0.089$ & $0.068$ & - & $0.308$ & $0.295$ \\
& ATT2IN~\cite{rennie2017self} & $0.224$ & $0.129$ & $0.089$ & $0.068$ & - & $0.308$ & $0.220$ \\
& HRGR~\cite{li2018hybrid}  & $0.438$ & $0.298$ & $0.208$ & $0.151$ & - & $0.322$ & $0.343$ \\
& COAT~\cite{jing2017automatic} & $0.455$ & $0.288$ & $0.205$ & $0.154$ & - & $0.369$ & $0.277$ \\
& R2GEN~\cite{chen2020generating}& $0.390$ & $0.293$ & $0.236$ & $0.161$ & $0.187$ & $0.371$ & - \\
& CMCL~\cite{liu2022competence} & $0.473$ & $0.305$ & $0.217$ & $0.162$ & $0.186$ & $0.378$ & - \\
& PPKED~\cite{liu2021exploring} & $0.483$ & $\colorbox{green!15}{\textbf{0.315}}$ & $0.224$ & $0.168$ & $0.190$ & $0.376$ & $0.351$ \\
& R2GENCMN~\cite{chen2022cross} & $0.402$ & $0.310$ & $0.225$ & $0.175$ & $0.190$ & $0.365$ & $0.344$ \\
& XPRONet~\cite{wang2022cross} & $0.480$ & $0.312$ & $0.225$ & $0.175$ & $0.190$ & $0.364$ & - \\
& RGRG~\cite{tanida2023interactive} & $0.373$ & $0.249$ & $0.175$ & $0.126$ & $0.168$ & $0.264$ & - \\
& M2KT~\cite{yang2023radiology} & $0.489$ & $\colorbox{green!15}{\textbf{0.315}}$ & $0.228$ & $0.174$ & - & $\colorbox{blue!15}{\textbf{0.379}}$ & $\colorbox{green!15}{\textbf{0.405}}$ \\
& GSKET~\cite{yang2022knowledge} & $\colorbox{green!15}{\textbf{0.490}}$ & $0.311$ & $0.236$ & $0.177$ & - & $0.367$ & $0.379$ \\
& COMG~\cite{gu2024complex} & $0.481$ & $0.313$ & $0.232$ & $0.183$ & $0.189$ & $0.371$ & - \\
&EKAGen~\cite{bu2024instance} & $0.471$ & $0.308$ & $0.223$ & $0.172$ & $0.201$ & $0.368$ & - \\
& CAMANET~\cite{wang2024camanet} & $0.478$ & $0.314$ & $0.229$ & $0.179$ & $\colorbox{green!15}{\textbf{0.202}}$ & $0.361$ & $0.412$ \\
& PromptMRG~\cite{jin2024promptmrg} &$0.488$  &$0.253$  &$\colorbox{green!15}{\textbf{0.237}}$  &$\colorbox{green!15}{\textbf{0.189}}$  & $0.192$ & $0.374$ & - \\
\cmidrule(l){2-9}
& RIHA (\textbf{Ours}) & $\colorbox{blue!15}{\textbf{0.498}}$ & $\colorbox{blue!15}{\textbf{0.317}}$ & $\colorbox{blue!15}{\textbf{0.239}}$ & $\colorbox{blue!15}{\textbf{0.218}}$& $\colorbox{blue!15}{\textbf{0.204}}$ & $\colorbox{green!15}{\textbf{0.376}}$& $\colorbox{blue!15}{\textbf{0.421}}$\\
\midrule
\multirow{10}{*}[-3pt]{MIMIC-CXR~\cite{johnson2019mimic}} & ST~\cite{vinyals2015show}  & $0.299$ & $0.184$ & $0.121$ & $0.084$ & $0.124$ & $0.263$ & - \\
& ATT2IN~\cite{rennie2017self} & $0.325$ & $0.203$ & $0.136$ & $0.096$ & $0.134$ & $0.276$ & - \\
& ADAATT~\cite{lu2017knowing} & $0.299$ & $0.185$ & $0.124$ & $0.088$ & $0.118$ & $0.266$ & - \\
& TOPDOWN~\cite{anderson2018bottom} & $0.317$ & $0.195$ & $0.130$ & $0.092$ & $0.128$ & $0.267$ & - \\
& R2GEN~\cite{chen2020generating} & $0.353$ & $0.218$ & $0.145$ & $0.103$ & $0.142$ & $0.270$ & - \\
& CMCL~\cite{liu2022competence} & $0.344$ & $0.217$ & $0.140$ & $0.097$ & $0.133$ & $0.281$ & - \\
& R2GENCMN~\cite{chen2022cross} & $0.348$ & $0.206$ & $0.135$ & $0.094$ & $0.136$ & $0.266$ & $0.158$ \\
& PPKED~\cite{liu2021exploring} & $0.360$ & $0.224$ & $0.149$ & $0.106$ & $\colorbox{green!15}{\textbf{0.149}}$ & $\colorbox{green!15}{\textbf{0.284}}$ & $\colorbox{blue!15}{\textbf{0.237}}$ \\
& XPRONet~\cite{wang2022cross} & $0.344$ & $0.215$ & $0.146$ & $0.105$ & $0.138$ & $0.279$ & $0.154$ \\
& M2KT~\cite{yang2023radiology} & $0.368$ & $0.220$ & $0.149$ & $\colorbox{green!15}{\textbf{0.107}}$ & - & $0.259$ & $0.110$ \\
&COMG~\cite{gu2024complex} & $0.345$ & $0.214$ & $0.143$ & $0.104$ & $0.136$ & $0.275$ & - \\
&EKAGen~\cite{bu2024instance} & $0.365$ & $\colorbox{green!15}{\textbf{0.233}}$ & $\colorbox{green!15}{\textbf{0.152}}$ & $0.105$ & $0.143$ & $0.271$ & - \\
& CAMANET~\cite{wang2024camanet} & $0.367$ & $0.219$ & $0.151$ & $\colorbox{green!15}{\textbf{0.107}}$ & $0.142$ & $0.279$ & $0.158$ \\
& PromptMRG~\cite{jin2024promptmrg} & $\colorbox{green!15}{\textbf{0.369}}$ &$0.223$  &$0.142$  &$0.099$  &$0.147$  &$0.260$  & - \\
\cmidrule(l){2-9}
 & RIHA (\textbf{Ours})  & $\colorbox{blue!15}{\textbf{0.385}}$ & $\colorbox{blue!15}{\textbf{0.251}}$ & $\colorbox{blue!15}{\textbf{0.162}}$ & $\colorbox{blue!15}{\textbf{0.109}}$ & $\colorbox{blue!15}{\textbf{0.159}}$ & $\colorbox{blue!15}{\textbf{0.293}}$ & $\colorbox{green!15}{\textbf{0.161}}$ \\
\bottomrule
\end{tabular}
\end{center}
\vspace{-15pt}
\end{table*}

\begin{table}[ht]
    \renewcommand\arraystretch{1.4}
    \setlength{\tabcolsep}{10.0pt}
    \centering
    \caption{Clinical efficacy comparisons on MIMIC-CXR dataset. \colorbox{blue!15}{Optimal} performance is highlighted. 
    \label{tab:efficacy}}
    \footnotesize  
    \begin{center}
    \begin{tabular}{@{}l c c c @{}}
        \toprule
        \textbf{Method}& \textbf{Precision} & \textbf{Recall} & \textbf{F1-Score}  \\ \midrule [\heavyrulewidth]
        ATT2IN~\cite{rennie2017self} & $0.268$ & $0.203$ &$0.204$   \\ 
        ADDAATT~\cite{lu2017knowing} & $0.322$ & $0.239$ &$0.249$   \\ 
        R2Gen~\cite{chen2020generating} & $0.406$ & $0.213$ & $0.280$   \\
        R2GenCMN~\cite{chen2022cross} & $0.440$  &$0.325$  &$0.374$  \\
        XProNet~\cite{wang2022cross} &  $0.463$& $0.285$ & $0.353$  \\
        PromptMRG~\cite{jin2024promptmrg} & $0.471$ & $0.198$ &$0.352$   \\
        RIHA(\textbf{Ours}) & $\colorbox{blue!15}{\textbf{0.486}}$ & $\colorbox{blue!15}{\textbf{0.298}}$ &$\colorbox{blue!15}{\textbf{0.375}}$   \\
      
        \bottomrule
    \end{tabular}
    \end{center}
    \vspace{-15pt}
\end{table}

\begin{table*}
\caption{The experimental results of ablation studies on the IU-Xray and MIMIC-CXR datasets.\colorbox{blue!15}{Optimal} performance is highlighted.}
\setlength{\fboxsep}{0.8pt}
\centering
\label{tab:ablation_studies}
\begin{tabular}{l|ccccccc}
\toprule  
\textbf{IU-Xray}  & \textbf{BLEU-1} & \textbf{BLEU-2} & \textbf{BLEU-3} &
\textbf{BLEU-4}  &\textbf{METEOR} & \textbf{ROUGE-L} & \textbf{CIDER}\\
\midrule  

Base  &0.390 &0.237 &0.161 &0.113 &0.170 & 0.322 &0.336 \\
+VFP  &0.419 &0.262 &0.184 &0.137 &0.172 & 0.340 &0.415 \\
+VFP+TFP+CHA &0.492 &0.313 &0.231 &$0.178$ &0.204 &$0.369$ &0.418 \\
+VFP+TFP+CHA+RPE &$\colorbox{blue!15}{\textbf{0.498}}$ &$\colorbox{blue!15}{\textbf{0.317}} $&$\colorbox{blue!15}{\textbf{0.239}}$ &\colorbox{blue!15}{\textbf{0.218}} &$\colorbox{blue!15}{\textbf{0.204}}$ &\colorbox{blue!15}{\textbf{0.376}} &$\colorbox{blue!15}{\textbf{0.421}} $\\

\toprule  
\textbf{MIMIC-CXR}  & \textbf{BLEU-1} & \textbf{BLEU-2} & \textbf{BLEU-3} &
\textbf{BLEU-4}  &\textbf{METEOR} & \textbf{ROUGE-L} & \textbf{CIDER} \\
\midrule  

Base  &0.363 &0.225 &0.152 &0.101 &0.142 & 0.280 &0.144\\
+VFP  &0.371  & 0.226 &0.155  &0.106  &0.152  &0.286   &0.150 \\
+VFP+TFP+CHA  &0.382  & 0.248 &0.158  &0.108  &0.158 &0.292  & 0.159 \\
+VFP+TFP+CHA+RPE  & $\colorbox{blue!15}{\textbf{0.385}}$ &$\colorbox{blue!15}{\textbf{0.251}}$  &$\colorbox{blue!15}{\textbf{0.162}}$  &$\colorbox{blue!15}{\textbf{0.109}}$  &$\colorbox{blue!15}{\textbf{0.159}}$ &$\colorbox{blue!15}{\textbf{0.293}}$  &$\colorbox{blue!15}{\textbf{0.161}}$  \\

\bottomrule 
\end{tabular}
\vspace{-15pt}
\end{table*}

\vspace{-5pt}
\subsection{Clinical Efficacy}
To assess the effectiveness of our proposed RIHA method in capturing abnormal clinical information, we compare its performance with several prior methods on the MIMIC-CXR dataset. The Clinical Efficacy (CE) metrics used in this evaluation were first introduced in R2Gen~\cite{chen2020generating}. For MIMIC-CXR, medical term extraction relies on the rule-based CheXpert labeler, which automatically generates labels for both reference and generated reports. Since the IU-Xray dataset lacks standardized labels, CE metrics are reported exclusively for MIMIC-CXR. As presented in Tab.~\ref{tab:efficacy}, our RIHA approach achieves the highest F1-Score of 0.375, surpassing R2GenCMN~\cite{chen2022cross} (0.374) and significantly outperforming XProNet (0.353) and PromptMRG~\cite{jin2024promptmrg} (0.352). In terms of precision, RIHA achieves 0.486, outperforming all baselines, including PromptMRG (0.471) and XProNet~\cite{wang2022cross} (0.463). For recall, RIHA attains 0.298, outperforming XProNet (0.285) and PromptMRG (0.198), though it falls short of R2GenCMN's recall (0.325). These results highlight RIHA's ability to balance precision and recall, demonstrating its potential for accurate medical condition identification in clinical practice.

\begin{table}[ht]
    \renewcommand\arraystretch{1.4}
    \setlength{\tabcolsep}{6pt}
    \centering
    \caption{Model performance under different hyperparameters $(\alpha, \gamma, \beta)$ on the IU-Xray dataset. \colorbox{blue!15}{Optimal} performance is highlighted.
    \label{tab:hyper}}
    \vspace{-5pt}
    \footnotesize  
    \begin{center}
    \begin{tabular}{@{}l c c c c @{}}
        \toprule
        \textbf{$(\alpha, \gamma, \beta)$}& \textbf{BLEU-1}  & \textbf{BLEU-4} & \textbf{METEOR} & \textbf{ROUGE-L}  \\ \midrule [\heavyrulewidth]

        (0.3, 0.4, 0.3)
        &$0.437$  &$0.162$  &$0.187$  &$0.340$   \\
        (0.4, 0.2, 0.4)  &$0.483$  &$0.172$  &$0.201$  &$0.362$  \\
        (0.5, 0.2, 0.3) & $ \colorbox{blue!15}{\textbf{0.498}}$  &$\colorbox{blue!15}{\textbf{0.218}}$  &$\colorbox{blue!15}{\textbf{0.204}}$  &$\colorbox{blue!15}{\textbf{0.376}}$   \\ 
        (0.6, 0.2, 0.2) &$0.471$  &$0.169$  &$0.198$  &$0.361$    \\ 
        (0.7, 0.2, 0.1) &$0.468$  &$0.165$ &$0.191$  &$0.359$   \\ 
        (1.0, 0.0, 0.0) & $0.449$ & $0.160$ &$0.181$  &  $0.354$ \\ 
        (0.0, 1.0, 0.0) & $0.429$ & $0.151$ &$0.173$  &$0.347$   \\
        (0.0, 0.0, 1.0) & $0.425$ &$0.136$  &$0.170$  &$0.343$    \\
        \bottomrule
    \end{tabular}
    \end{center}
    \vspace{-15pt}
\end{table}

\begin{table}[ht]
    \renewcommand\arraystretch{1.4}
    \setlength{\tabcolsep}{3.0pt}
    \centering
    \caption{Effect of using different BERT pre-trained models to extract the textual features on the IU-Xray dataset. \colorbox{blue!15}{Optimal} performance is highlighted.
    \label{tab:BERTmodel}}
    \vspace{-5pt}
    \footnotesize  
    \begin{center}
    \begin{tabular}{@{}l c c c c @{}}
        \toprule
        \textbf{BERT-type}&\textbf{ BLEU-1} &\textbf{ BLEU-4} & \textbf{METEOR} & \textbf{ROUGE-L} \\ \midrule [\heavyrulewidth]

        bert-base-uncased
        &$0.487$  & $0.198$ & $0.196$ &$0.349$   \\
        PubMedBERT-base & $0.496$ & $0.213$ &$0.204$  &$0.371$   \\ 
        Bio\_ClinicalBERT & $\colorbox{blue!15}{\textbf{0.498}}$ &$\colorbox{blue!15}{\textbf{0.218}}$  &$\colorbox{blue!15}{\textbf{0.204}}$  &$\colorbox{blue!15}{\textbf{0.376}}$  \\
        
        \bottomrule
    \end{tabular}
    \end{center}
    \vspace{-15pt}
\end{table}

\begin{table}[ht]
    \renewcommand\arraystretch{1.4}
    \setlength{\tabcolsep}{5pt}
    \centering
    \caption{Effect of using different distance calculation methods for the proposed CHA module on IU-Xray dataset. \colorbox{blue!15}{Optimal} performance is highlighted.
    \label{tab:distance}}
    \vspace{-5pt}
    \footnotesize  
    \begin{center}
    \begin{tabular}{@{}l c c c c @{}}
        \toprule
        \textbf{Distance Calculation}& \textbf{BLEU-1} & \textbf{BLEU-4} & \textbf{METEOR} & \textbf{ROUGE-L}  \\ \midrule [\heavyrulewidth]

        Wasserstein distance
        &$\colorbox{blue!15}{\textbf{0.498}}$  &$\colorbox{blue!15}{\textbf{0.218}}$  &$\colorbox{blue!15}{\textbf{0.204}}$  &$\colorbox{blue!15}{\textbf{0.376}}$    \\
        
        $L_{1}$ distance & $0.487$ &$0.209$  &$0.199$  &$0.368$   \\
        $L_{2}$ distance &$0.482$  &$0.205$  &$0.197$  &$0.364$    \\ 
        Cosine distance &$0.491$  &$0.213$  &$0.201$  &$0.371$   \\ 
        KL divergence &$0.476$  &$0.201$  &$0.195$  &$0.360$   \\ 
        \bottomrule
    \end{tabular}
    \end{center}
    \vspace{-15pt}
\end{table}

\vspace{-5pt}
\subsection{Ablation Studies}
In this section, we evaluate the effectiveness of our proposed modules, including the visual feature pyramid (VFP), cross-modal hierarchical alignment (CHA), and relative positional embedding (RPE). We also provide a detailed analysis of hyperparameter settings for hierarchical alignment loss, the choice of textual pre-trained models, and distance calculation methods used in the alignment loss.

\subsubsection{Effectiveness of proposed modules}
The following models are evaluated in our experiment.
\begin{itemize}
\item \textbf{Base}: The base model only includes the basic visual extractor and the basic transformer encoder-decoder structure.
\item \textbf{Base+VFP}: The visual features are extracted from the $5$-th, $6$-th and $7$-th layer of the basic visual extractor, and then are fused to the subsequent module.
\item \textbf{Base+VFP+TFP+CHA}: This introduces the multi-granularity alignment by the CHA module between visual features from different layers and textual features from paragraph, sentence, and word levels.
\item \textbf{Base+VFP+TFP+CHA+RPE}: This is the full RIHA containing all our proposed components.
\end{itemize}
As shown in Tab.~\ref{tab:ablation_studies}, we progressively integrated three critical components into our base model: Visual Feature Projection (VFP), Cross-modal Hierarchical Attention (CHA), and Relative Positional Embedding (RPE). Each addition demonstrated notable performance gains across multiple metrics. On the IU-Xray dataset, incorporating VFP alone increased the BLEU-1 score from 0.390 to 0.419. Adding CHA further boosted scores (BLEU-1: 0.492, BLEU-4: 0.178), while the full model, including RPE, achieved the highest performance (BLEU-1: 0.498, CIDEr: 0.421, METEOR: 0.207). Similar improvements were observed on the more challenging MIMIC-CXR dataset, where each component incrementally enhanced the overall results, culminating in the best scores for the complete model (BLEU-1: 0.385, ROUGE-L: 0.293). These results underscore the distinct and complementary contributions of each component to overall report generation quality.

\subsubsection{Analysis on hyperparameters for alignment loss}
We conduct extensive experiments to analyze the sensitivity of the hyper-parameters $\alpha$, $\gamma$, and $\beta$ at the paragraph, sentence, and word levels, respectively, across various evaluation metrics (see Tab.~\ref{tab:hyper}). The results show that the configuration (0.5, 0.2, 0.3) achieves the best performance, with BLEU-1, BLEU-4, METEOR, and ROUGE-L scores of 0.498, 0.176, 0.207, and 0.368, respectively. We observe that increasing the paragraph-level weight ($\alpha$) from 0.3 to 0.5 consistently improves performance, highlighting the importance of paragraph-level coherence. However, further increasing $\alpha$ beyond 0.5 leads to diminishing returns, indicating a trade-off between paragraph-level emphasis and sentence/word-level features. Extreme cases where one component is maximized (e.g., $\alpha=1.0$, $\gamma=1.0$, or $\beta=1.0$) result in significantly lower performance, confirming that multi-level integration is crucial for accurate text evaluation. This supports our hypothesis that focusing exclusively on any single linguistic level overlooks the complex interdependencies that drive overall text quality.

\subsubsection{Analysis on textual pre-trained models}
To extract multi-granularity textual features, we use the pre-trained BERT model, which remains frozen during training. While the base BERT model is trained on natural language, Bio\_ClinicalBERT and PubMedBERT are fine-tuned on clinical text to capture domain-specific medical language patterns. Intuitively, domain-specific BERT models should yield better results. Tab.~\ref{tab:BERTmodel} compares the performance of different BERT architectures on the IU-Xray dataset. The results show a clear advantage for domain-specific models. The general-purpose bert-base-uncased model performs reasonably well (BLEU-1: 0.487, BLEU-4: 0.198), but domain-specific models show significant improvements across all metrics. PubMedBERT-base, pre-trained on biomedical literature, improves BLEU-4 (0.213) and ROUGE-L (0.371) scores. Notably, Bio\_ClinicalBERT, pre-trained on clinical text, outperforms all models (BLEU-1: 0.498, BLEU-4: 0.218, METEOR: 0.204, ROUGE-L: 0.376). These findings confirm that domain-specific pretraining effectively captures medical terminology and semantic relationships, improving the accuracy and relevance of radiology report generation.

\subsubsection{Analysis on the distance calculation}
To assess the impact of different distance metrics on the alignment of visual and textual data distributions, we conduct experiments on the IU-Xray dataset using various distances: Wasserstein, $L_{1}$, $L_{2}$, Cosine, and KL divergence. The results, shown in Tab.~\ref{tab:distance}, indicate that Wasserstein distance consistently outperforms the others, achieving the highest scores across all metrics (BLEU-1: 0.498, BLEU-4: 0.218, METEOR: 0.204, ROUGE-L: 0.376). This superior performance is due to the Wasserstein distance's ability to capture geometric relationships between distributions, even with minimal overlap. Cosine distance also performs competitively (BLEU-1: 0.491, ROUGE-L: 0.371), highlighting that directional similarity effectively captures cross-modal relationships. Traditional $L_{1}$ and $L_{2}$ distances produce moderate results, with $L_{1}$ consistently outperforming $L_{2}$, suggesting that the Manhattan distance better suits the discrete nature of our task. KL divergence shows the lowest performance, likely due to its sensitivity to zero-probability regions and its asymmetric nature. These findings indicate that optimal cross-modal alignment is best achieved by using distance metrics that consider the structural properties of underlying distributions, rather than simply comparing probability masses.

\vspace{5pt}
\subsubsection{Analysis on the computational cost of distance calculation methods}Due to the high computational cost of the Wasserstein distance, we conduct experiments to evaluate the computational cost between standard OT calculation and existing other optimized OT calculations in Tab.~\ref{tab:OT-strategy}. Based on the table, the results demonstrate that both accelerated OT strategies substantially improve computational efficiency. Sampling-based OT achieves the greatest reduction with 43\% faster training and 24\% lower memory consumption, while prototype-based OT provides a 32\% speedup in training time and 20\% reduction in GPU memory usage compared to standard OT. In this work, we employed standard OT to thoroughly investigate its suitability and effectiveness for hierarchical alignment in radiology report generation. Future work will explore how optimized OT solvers can further enhance both computational efficiency and model performance.

\subsubsection{Analysis on the backbone of visual extractor}To validate our choice of ResNet101 as the visual backbone, we conduct comparative experiments with alternative architectures on the IU-Xray dataset. As shown in table~\ref{tab:visualbackbone}, ResNet101 achieves the best performance across all metrics, with BLEU-1 of 0.498, BLEU-4 of 0.218, METEOR of 0.204, and ROUGE-L of 0.376. DenseNet121 demonstrates competitive but slightly lower performance (BLEU-1: 0.485, BLEU-4: 0.206), while ViT shows the weakest results (BLEU-1: 0.479, BLEU-4: 0.201). The superior performance of ResNet101 can be attributed to its robust feature extraction capability for medical images and well-established pre-training on large-scale datasets. Given the relatively small size of medical datasets, CNN-based architectures like ResNet101 demonstrate better generalization than transformer-based models, which typically require larger training corpora. These results justify our architectural choice and align with previous findings in the radiology report generation literature.

\subsubsection{Analysis on the frozen and finetuned BERT encoder} To investigate the impact of BERT encoder training strategy on report generation performance, we compared frozen and fine-tuned Bio\_ClinicalBERT encoders on the IU-Xray dataset. As shown in the table, the frozen BERT encoder achieves comparable or slightly better performance across all metrics (BLEU-1: 0.498, BLEU-4: 0.218, METEOR: 0.204, ROUGE-L: 0.376) compared to the fine-tuned version (BLEU-1: 0.495, BLEU-4: 0.215, METEOR: 0.202, ROUGE-L: 0.373). The marginal performance difference (0.5-1.5\% across metrics) suggests that fine-tuning the pre-trained BERT encoder provides minimal additional benefit for our hierarchical alignment framework. This result can be attributed to several factors. First, Bio\_ClinicalBERT has already been extensively pre-trained on clinical text, capturing domain-specific medical terminology and semantic relationships effectively. Second, freezing the BERT encoder prevents potential overfitting and forgetting to the relatively small training corpus and preserves the rich linguistic knowledge learned during pre-training. Based on these findings, we adopt the frozen BERT encoder strategy in our final model, which offers the advantages of reduced computational cost, faster training convergence, while maintaining competitive performance.

\begin{table}[ht]
    \renewcommand\arraystretch{1.4}
    \setlength{\tabcolsep}{3.0pt}
    \centering
    \caption{Effect of using different frozen or fine-tuned BERT to extract the textual features on the IU-Xray dataset. 
    \label{tab:frozen}}
    \vspace{-5pt}
    \footnotesize  
    \begin{center}
    \begin{tabular}{@{}l c c c c @{}}
        \toprule
        \textbf{BERT encoder}&\textbf{ BLEU-1} &\textbf{ BLEU-4} & \textbf{METEOR} & \textbf{ROUGE-L} \\ \midrule [\heavyrulewidth]
        Frozen
        &$0.498$  & $0.218$ & $0.204$ &$0.376$   \\
        Fine-tuned & $0.495$ & $0.215$ &$0.202$  &$0.373$   \\ 
        
        \bottomrule
    \end{tabular}
    \end{center}
    \vspace{-15pt}
\end{table}

\begin{table}[ht]
    \renewcommand\arraystretch{1.4}
    \setlength{\tabcolsep}{3.0pt}
    \centering
    \caption{Effect of using different visual encoders to extract the visual features on the IU-Xray dataset. \colorbox{blue!15}{Optimal} performance is highlighted.
    \label{tab:visualbackbone}}
    \vspace{-10pt}
    \footnotesize  
    \begin{center}
    \begin{tabular}{@{}l c c c c @{}}
        \toprule
        \textbf{Visual Encoder}&\textbf{ BLEU-1} &\textbf{ BLEU-4} & \textbf{METEOR} & \textbf{ROUGE-L} \\ \midrule [\heavyrulewidth]

        ResNet101
        &$\colorbox{blue!15}{\textbf{0.498}}$  & $\colorbox{blue!15}{\textbf{0.218}}$ & $\colorbox{blue!15}{\textbf{0.204}}$ &$\colorbox{blue!15}{\textbf{0.376}}$   \\
        DenseNet121 & $0.485$ & $0.206$ &$0.197$  &$0.368$   \\ 
        ViT & $0.479$ &$0.201$  &$0.193$  &$0.362$  \\
        
        \bottomrule
    \end{tabular}
    \end{center}
    \vspace{-15pt}
\end{table}

\begin{table}[ht]
    \renewcommand\arraystretch{1.4}
    \setlength{\tabcolsep}{3.0pt}
    \centering
    \caption{Computational cost of different distance calculation methods on the IU-Xray dataset. 
    \label{tab:OT-strategy}}
    \vspace{-8pt}
    \footnotesize  
    \begin{center}
     \resizebox{\linewidth}{!}{
    \begin{tabular}{@{}l c c  @{}}
        \toprule
        \textbf{Distance Strategy}&\textbf{Training Time(h)} &\textbf{GPU Memory Usage(GB)}  \\ \midrule [\heavyrulewidth]
        $L_{1}$ Distance &$1.2$  & $8.4$    \\
        $L_{2}$ Distance &$1.3$  & $8.6$    \\
        Cosine Similarity &$1.4$  & $8.8$    \\
        Standard OT &$2.8$  & $12.1$    \\
        Sampling-based OT & $1.6$ & $9.2$   \\ 
        Prototype-based OT & $1.9$ &$9.7$    \\
        
        \bottomrule
    \end{tabular}}
    \end{center}
    \vspace{-15pt}
\end{table}

\begin{figure*}
    \centering 
    \includegraphics[width=0.95\textwidth]{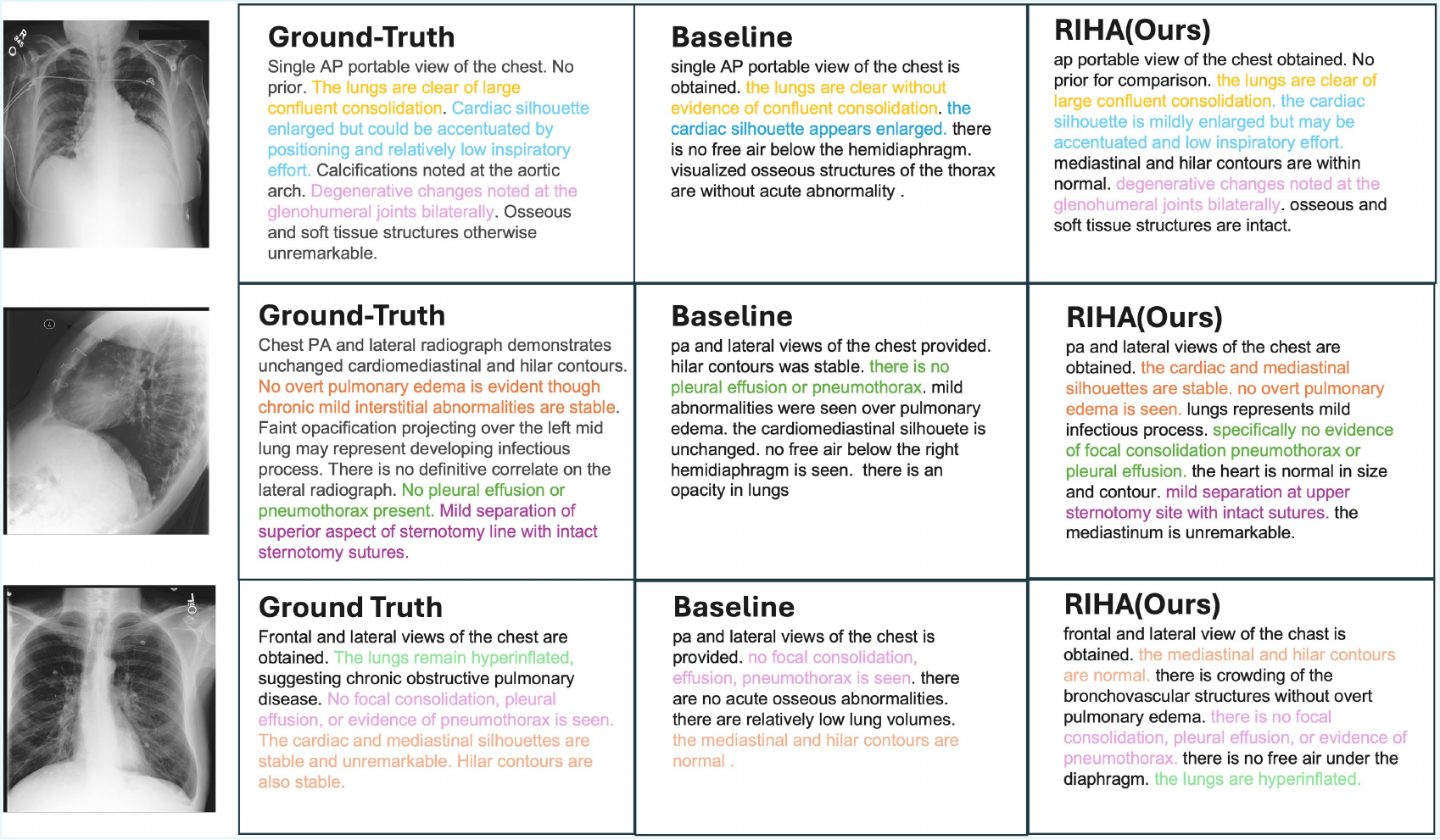} 
    \caption{Examples of generated reports from the MIMIC-CXR testing subset using the baseline model and our proposed RIHA method. Identical findings in the ground truth (GT) and generated reports are highlighted with matching colors, demonstrating the superior performance of our approach.}    
    \label{report}  
    \vspace{-15pt}
\end{figure*}

\begin{figure}
    \centering 
    \includegraphics[width=0.48\textwidth]{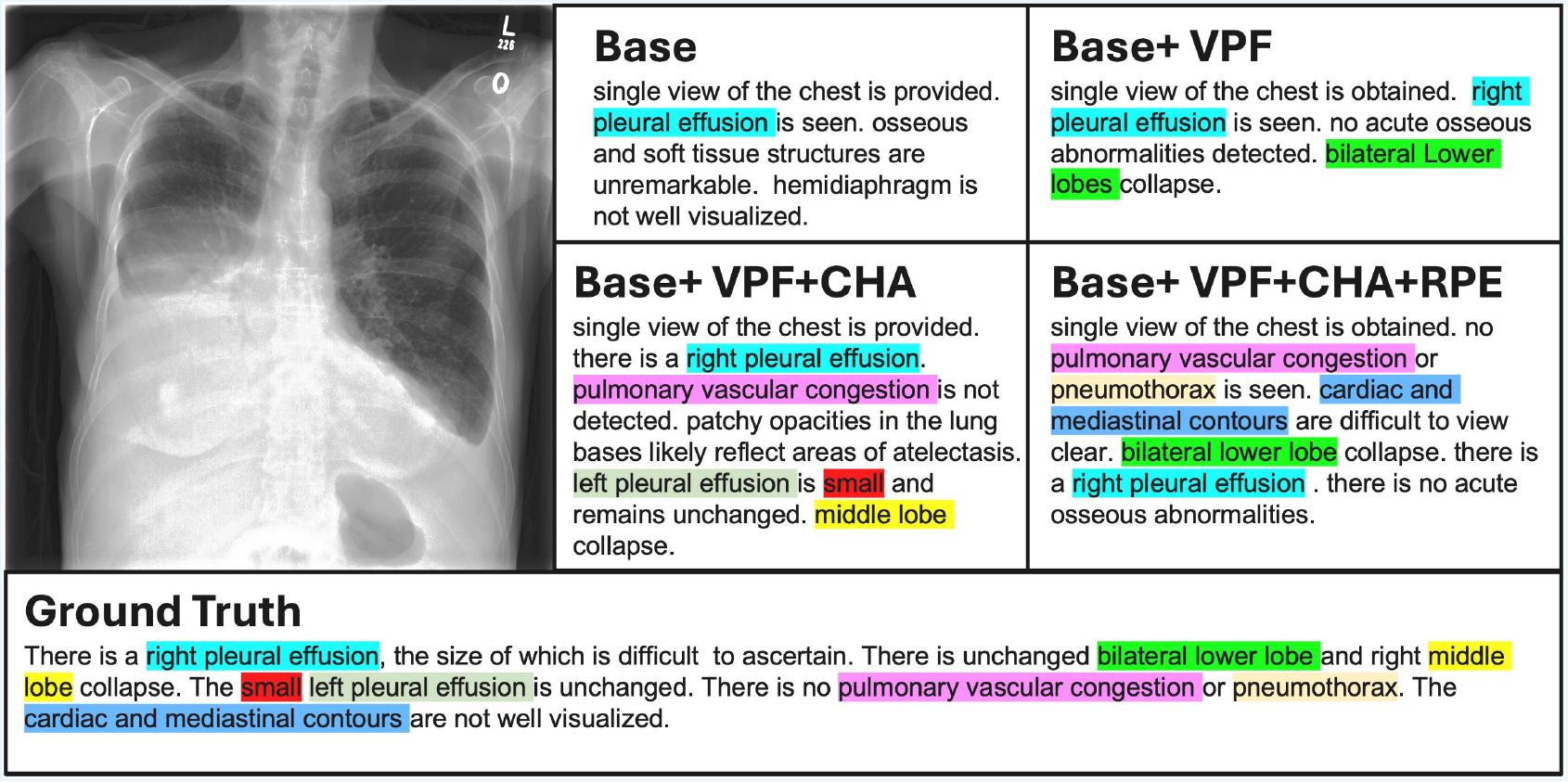} 
    \caption{Example reports generated by incrementally incorporating the proposed modules into the baseline model are shown. Key medical terms are highlighted in different colors to clearly differentiate model performance.}   
    \label{fig:ab-vis}  
    \vspace{-10pt}
\end{figure}

\vspace{-5pt}
\subsection{Qulitative Results}

\begin{figure}
    \centering 
    \includegraphics[width=0.5\textwidth]{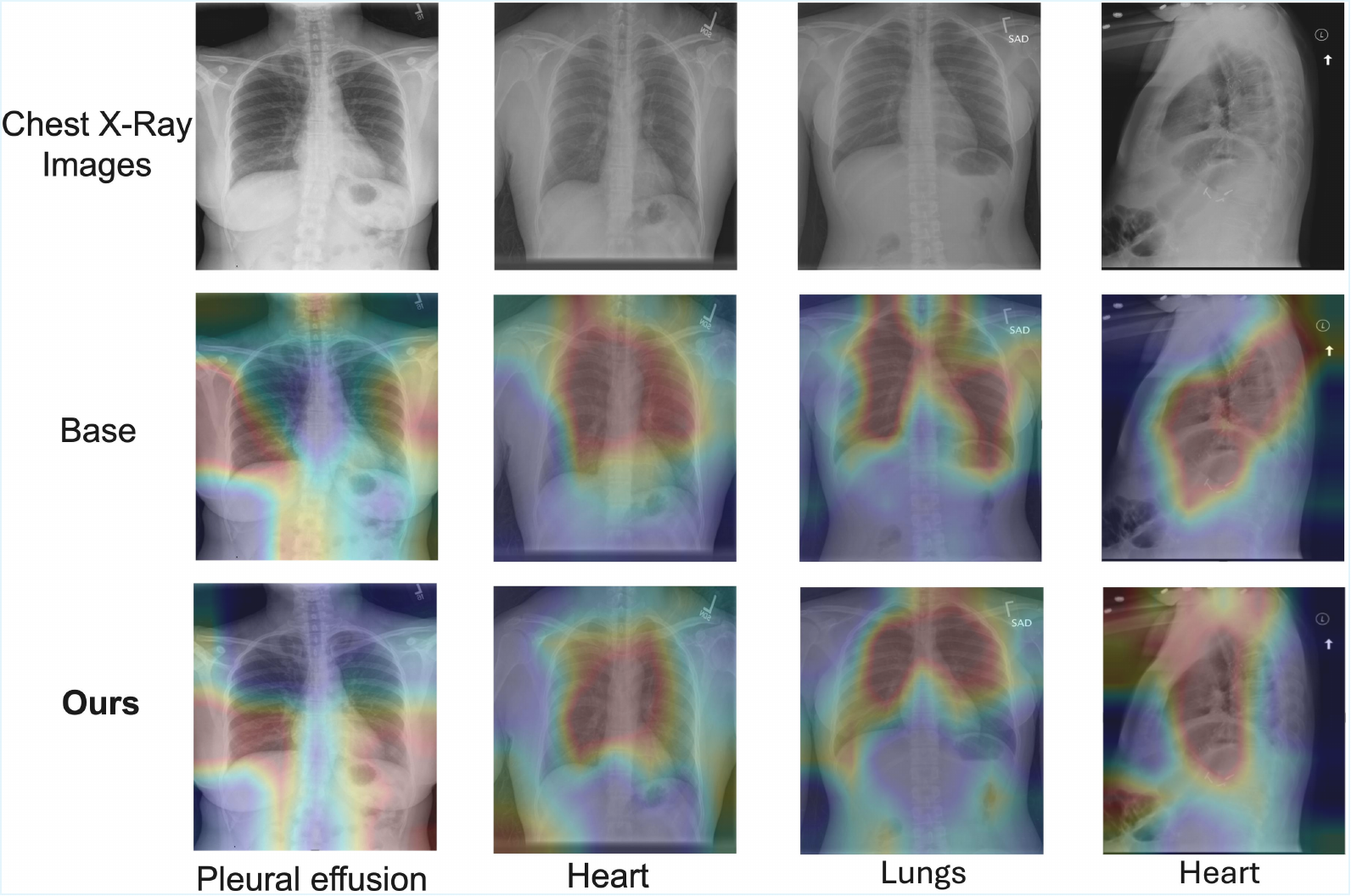} 
    \caption{Attention map visualizations for various keywords from the Baseline and RIHA models reveal that RIHA assigns more precise attention regions, highlighting its improved focus for each keyword.}
    \label{fig:attmap} 
    \vspace{-8pt}
\end{figure}

To further assess the effectiveness of our proposed method, we present qualitative results in Fig.~\ref{report}, Fig.~\ref{fig:ab-vis}, and Fig.~\ref{fig:attmap}. Fig.~\ref{report} highlights the superior performance of our RIHA method in generating radiology reports. Unlike the baseline model, which incorrectly identifies ``confluent consolidation" in the anteroposterior (AP) view, RIHA accurately detects ``clear of large confluent consolidation" and appropriately evaluates the cardiac silhouette, aligning closely with ground-truth annotations. Similarly, in posteroanterior (PA) and lateral views, RIHA correctly identifies the absence of pleural effusion and reports ``mild infectious process, specifically no evidence of focal consolidation pneumothorax or pleural effusion," reflecting a more precise assessment. Fig.~\ref{fig:ab-vis} further demonstrates the incremental improvements from our architectural components. The base model with Visual Feature Pyramid (VFP) extractor shows enhanced detection of bilateral lower lobe collapse, while adding Cross-Hierarchy Attention (CHA) enables more accurate identification of pulmonary vascular congestion and atelectasis. The complete model (Base+VFP+CHA+RPE) achieves comprehensive analysis, accurately detecting pleural effusions and providing clearer evaluations of cardiac and mediastinal contours. Finally, Fig.~\ref{fig:attmap} presents attention maps, illustrating our model's capacity to align textual descriptions with corresponding anatomical regions. These heat maps show precise localization of pleural effusion, cardiac structures, and lung fields, validating RIHA's ability to effectively learn the spatial relationships between radiological findings and their textual descriptions. This visual evidence supports our quantitative results, confirming RIHA's potential for generating expert-level radiological interpretations.

\subsection{Clinical Implications}
\subsubsection{Clinical analysis by LLM and radiologists}To assess the clinical interpretability and quality of generated reports, we conduct a human expert evaluation study. We randomly sample 30 cases from the test set and generate reports using Baseline, R2Gen, XProNet, and our proposed RIHA model. A board-certified radiological expert from Tan Tock Seng Hospital (TTSH) independently evaluated all generated reports across multiple dimensions, including clinical correctness, coherence, completeness, and diagnostic accuracy, providing an overall score for each model. As shown in the Tab.~\ref{tab:score}, RIHA achieves the highest clinician score of 8.0/10, substantially outperforming XProNet (6.7), R2Gen (6.3), and the Baseline model (5.8). Notably, RIHA's score demonstrates the closest alignment between automated LLM evaluation (8.2) and clinical expert judgment, with a minimal gap of only 0.2 points, compared to larger discrepancies observed in other models (0.4-0.5 points). This strong concordance indicates that RIHA-generated reports better capture the clinical reasoning patterns and diagnostic standards expected by a radiological expert.
\vspace{5pt}
\subsubsection{Clinical error analysis}

To evaluate the clinical reliability of our approach beyond standard NLG metrics, we conduct a comprehensive analysis of critical finding detection accuracy. Using the CheXpert~\cite{irvin2019chexpert} labeler to automatically annotate both GT and generated reports, we perform a focused case study on cardiomegaly detection in the IU-Xray test set - a critical cardiac finding that significantly impacts patient management decisions. We categorize errors into three types: Omission (O), where a finding present in GT is absent from the generated report; Misclassification (M), where GT and generated reports have conflicting conclusions; and Hallucination (H), where a finding appears in the generated report but not in GT. Based on the definitions, M encompasses both O and H, i.e., M = O + H. As shown in Tab.~\ref{tab:clinical-error}, RIHA achieves superior clinical accuracy with a $94.8\%$ detection rate and only $5.2\%$ misclassification rate for cardiomegaly cases, comprising $3.1\%$ omission and $2.1\%$ hallucination errors. This represents a substantial improvement over baseline methods ($79\%$ detection, $21\%$ misclassification) and SOTA approaches including R2GenCMN ($81\%$ detection, $19\%$ misclassification) and XProNet ($86.2\%$ detection, $13.8\%$ misclassification). Notably, RIHA demonstrates the lowest omission rate ($3.1\%$) among all methods, indicating its effectiveness in capturing critical findings. The significantly reduced error rates across all categories demonstrate the clinical value of our hierarchical alignment approach in detecting subtle but clinically significant abnormalities.

\begin{table}[ht]
    \renewcommand\arraystretch{1.4}
    \setlength{\tabcolsep}{4.0pt}
    \centering
    \caption{Cardiomegaly detection and error rates by different models. The characters O, M, and  H represent Omission, Misclassification, and Hallucination, respectively.
    \label{tab:clinical-error}}
    \vspace{-5pt}
    \footnotesize  
    \begin{center}
    \begin{tabular}{@{}l c c c c @{}}
        \toprule
        \textbf{Model}&\textbf{Cardiomegaly detected} &\textbf{M} & \textbf{O} &\textbf{H}  \\ \midrule [\heavyrulewidth]

        RIHA &$94.8\%$  & $5.2\%$ & $3.1\%$ & $2.1\%$  \\
        Baseline & $79.0\%$ & $21.0\%$ & $14.5\%$ & $6.5\%$  \\ 
        R2GenCMN & $81.0\%$ & $19.0\%$ & $13.1\%$ & $5.9\%$  \\
        XProNet & $86.2\%$ & $13.8\%$  & $9.2\%$ & $4.6\%$  \\
        \bottomrule     
    \end{tabular}
    \end{center}
    \vspace{-15pt}
\end{table}

\begin{table}[ht]
    \renewcommand\arraystretch{1.4}
    \setlength{\tabcolsep}{3.0pt}
    \centering
    \caption{Evaluation score of the generated reports from different models by LLM and clinicians.
    \label{tab:score}}
    \vspace{-10pt}
    \footnotesize  
    \begin{center}
    \begin{tabular}{@{}l c c c c @{}}
        \toprule
        \textbf{Model}&\textbf{Baseline} &\textbf{R2Gen} & \textbf{XProNet} & \textbf{RIHA} \\ \midrule [\heavyrulewidth]

        LLM Score
        &$6.2$  & $6.8$ & $7.1$ &$8.2$   \\
        Clinician Score & $5.8$ & $6.3$ &$6.7$  &$8.0$   \\ 
        \bottomrule
    \end{tabular}
    \end{center}
    \vspace{-15pt}
\end{table}

\vspace{-5pt}
\section{Limitations and Discussions}
Despite its superior performance in radiology report generation, our current model has some limitations. First, it relies heavily on the manual disentanglement of three granularities, a step that could be streamlined in the future by leveraging large language models for automatic multi-granularity selection and disentanglement. We could train lightweight segmentation modules to identify paragraph and sentence boundaries by leveraging the consistent structure of radiology reports. Alternatively, we could utilize medical-specific large language models to automatically extract semantically meaningful keywords while maintaining clinical relevance. Second, the optimal transport algorithm used for alignment is computationally intensive, but this can be mitigated by enhancing the original algorithm, for instance, through optimized sampling or prototype-based refinement. Nonetheless, our model remains efficient and lightweight, with the hierarchical alignment module being removable during testing, significantly boosting inference speed. This modular design also makes the approach adaptable to related tasks, such as paragraph generation, text-to-image, or image-to-text generation. Third, our current framework handles uncertainty expressions (e.g., ``mild," ``possible," ``likely") implicitly through the pre-trained Bio\_ClinicalBERT encoder rather than explicitly modeling the graded nature of clinical uncertainty. Inspired by recent uncertainty-aware approaches~\cite{wangcurv, najdenkoska2022uncertainty}, future work could incorporate explicit uncertainty scores to weight cross-modal alignment, assigning stronger weights to definite findings while attenuating uncertain expressions. Finally, though RIHA's hierarchical alignment framework demonstrates strong theoretical potential for cross-modal adaptation, generalizability is limited due to our current evaluation focusing on chest X-rays. For CT scans, adaptation presents domain-specific challenges including slice interdependency modeling, variable slice thickness, contrast phase dependencies, and specialized preprocessing requirements for CT's wide dynamic range. MRI adaptation introduces complex considerations such as multi-sequence fusion for T1, T2, FLAIR sequences, sequence-specific pathology visibility, and motion artifact handling. While our modular pyramid-based architecture provides flexibility for different report formats, structured templates introduce challenges like template constraint integration for standardized systems and cross-institutional format variations. Despite these modality-specific challenges, our hierarchical alignment principles offer substantial adaptability potential, though practical implementation would require domain-specific modifications and extensive validation. However, accurately capturing rare and subtle findings remains a core challenge in medical report generation. To address this, future work could benefit from the robust in-context learning capabilities of advanced vision-language models to further refine performance.

\vspace{-5pt}
\section{Conclusion}
 In this work, we propose RIHA, a report-image hierarchical alignment model designed to achieve fine-grained alignment by hierarchically aligning textual and visual information at the paragraph, sentence, and word levels. The optimal transport principle is leveraged for the fine-grained alignment goal, and the relative positional embedding strategy helps enhance the alignment further at the token level. Extensive experiments on benchmark chest X-ray image-report datasets, including IU-Xray and MIMIC-CXR, demonstrate that this approach significantly improves the comprehensiveness and precision of generated radiology reports. These results indicate that a carefully designed fine-grained alignment network can effectively enhance the quality of cross-modal medical report generation.

\vspace{-5pt}
\section*{References}

\bibliographystyle{IEEEtran}
\bibliography{IEEEabrv,refs}

\end{document}